\documentclass[a4paper, 10pt, conference]{ieeeconf}      
\usepackage[noadjust]{cite}
\usepackage{FG2025}
\usepackage{url}
\usepackage{amsfonts}
\usepackage{amsmath}
\usepackage{graphicx}
\usepackage{makecell}
\usepackage{tabularx}
\usepackage{hyperref}

\FGfinalcopy 

\def\eg{\emph{e.g.}, }

\def\ie{\emph{i.e.}, }

\IEEEoverridecommandlockouts                              
\def\FGPaperID{184} 

\title{\LARGE \bf
Beyond FACS: Data-driven Facial Expression Dictionaries, \\with Application to Predicting Autism}

\author{\parbox{16cm}{\centering
    {\large Evangelos Sariyanidi,$^1$ Lisa Yankowitz,$^1$ Robert T. Schultz,$^{1,2}$ John D. Herrington,$^{1,2}$\\ Birkan Tunc$^{\dagger1,2}$ and Jeffrey Cohn$^{\dagger 3}$}\\
    {\normalsize
    $^1$ Center for Autism Research, The Children’s Hospital of Philadelphia, Philadelphia, PA, USA\\
    $^2$ Department of Psychiatry, University of Pennsylvania, Philadelphia, PA, USA\\
    $^3$ Deliberate AI, New York, NY, USA.\\
    $^\dagger$ Equal contribution.}}
}

\begin{document}

\ifFGfinal
\thispagestyle{empty}
\pagestyle{empty}
\else
\author{Anonymous FG2025 submission\\ Paper ID \FGPaperID \\}
\pagestyle{plain}
\fi
\maketitle


\begin{abstract}
The Facial Action Coding System (FACS) has been used by numerous studies to investigate the links between facial behavior and mental health. The laborious and costly process of FACS coding has motivated the development of machine learning frameworks for Action Unit (AU) detection. Despite intense efforts spanning three decades, the detection accuracy for many AUs is considered to be below the threshold needed for behavioral research. Also, many AUs are excluded altogether, making it impossible to fulfill the ultimate goal of FACS\----the representation of \textit{any} facial expression in its entirety. This paper considers an alternative approach. Instead of creating automated tools that mimic FACS experts, we propose to use a new coding system that mimics the key properties of FACS. Specifically, we construct a data-driven coding system called the Facial Basis, which contains units that correspond to localized and interpretable 3D facial movements, and overcomes three structural limitations of automated FACS coding. First, the proposed method is completely unsupervised, bypassing costly, laborious and variable manual annotation. Second, Facial Basis reconstructs all observable movement, rather than relying on a limited repertoire of recognizable movements (as in automated FACS). Finally, the Facial Basis units are additive, whereas AUs may fail detection when they appear in a non-additive combination. The proposed method outperforms the most frequently used AU detector in predicting autism diagnosis from in-person and remote conversations, highlighting the importance of encoding facial behavior comprehensively. To our knowledge, Facial Basis is the first alternative to FACS for deconstructing facial expressions in videos into localized movements. We provide an open source implementation of the method at \url{github.com/sariyanidi/FacialBasis}.  
\end{abstract}

\section{INTRODUCTION}

Since its initial development more than four decades ago~\cite{ekman78}, the Facial Action Coding System (FACS) has been widely used to study how facial expressions relate to emotion, personality, mood, deception, and mental health~\cite{ekman05}. With the advent of computer vision, researchers have aimed to automate the detection of FACS Action Units (AUs), as manual AU coding is laborious, costly, and requires extensive training~\cite{ekmanwebsite}. However, despite nearly three decades of research, the accuracy of automated systems is often below the threshold needed for behavioral research~\cite{ertugrul20}. Moreover, available AU detection software excludes a large number of AUs\----and even the AUs that are included may fail detection if they are annotated with low reliability or appear in a non-additive AU combination~(Section~\ref{sec:AU}). As such, automated FACS coders behave more akin to a retrieval system that checks for the presence of certain AUs or AU combinations, rather than fulfilling the original purpose of FACS, which is encoding any facial expression in its entirety. The consequences of this major goalpost shift remain unknown and are potentially severe for progress in behavioral and medical sciences. The inability of automated FACS coders to fully encode expressions can prevent the discovery of behavioral patterns that characterize the full range of emotions, personality traits or mental health conditions.

Instead of developing a new automated FACS coder, we propose a new, alternative coding system that is more amenable to automated expression measurement. Specifically, we propose to construct a data-driven coding system by learning a dictionary~\cite{mairal10}, which can overcome three significant inherent limitations of automated FACS coding. First, dictionaries reconstruct all observable facial movements, therefore can encode expressions comprehensively, in contrast with FACS software that provide results only for the AUs in their repertoire. Second, dictionaries are learned in an unsupervised manner. This is a significant advantage over FACS-based approaches, as the costly and laborious manual annotation needed by the latter may render large parts of available videos unusable for supervised training~(Section~\ref{sec:AU}). Third, all basic expression units in a dictionary are additive\----they can be detected successfully in isolation or in combination with other expression units~(Section~\ref{sec:AU}).


To show the utility of data-driven dictionaries in clinical applications, we compare multiple facial coding systems, including (automated) FACS coding, in classifying adolescents with autism (AUT) vs.\ those who are neurotypical (NT). We experiment on two datasets of naturalistic conversational tasks: one with in-person conversations, and one with remote conversations. Studying both contexts allows us to assess whether behavioral symptoms that characterize autism can be effectively measured from remote conversations, which is needed given the post-pandemic increase in telehealth assessment of autism~\cite{de23}. The present results suggest that the proposed system outperforms the most widely used automated FACS coder, namely OpenFace~\cite{baltrusaitis16}, in AUT vs.\ NT classification both on the in-person and the remote sample. Furthermore, results from remote conversations underpin the importance of comprehensively encoding facial expressions (Section~\ref{sec:results}), suggesting that the exclusion of many AUs from automated pipelines (Section~\ref{sec:AU}) may be consequential.

Our findings indicate that developing data-driven coding systems that retain the advantages of FACS, rather than automating FACS, is a potent new paradigm for empowering mental health research~(Section~\ref{sec:discussion}), since the structural limitations of automated FACS coding~(Section~\ref{sec:AU}) may not be overcome by developing more sophisticated AU detectors.  We provide an end-to-end open-source toolkit\footnote{\url{https://github.com/sariyanidi/FacialBasis}} that can be used for conducting behavioral research with the Facial Basis. To our knowledge, this is the first open-source software that provides an alternative to automated FACS for quantifying facial expressions in 2D videos by breaking them down into localized expression units.

In sum, the contributions of this paper are as follows. We:

\begin{itemize}
\item Show that data-driven dictionaries are a viable alternative to FACS for supporting mental health research with an interpretable expression coding system.
\item Experiment using in-person and remote conversations and show that autism can be predicted in both contexts.
\item Show that the behavioral symptoms that are most predictive of AUT vary between contexts.
\item Provide an open-source toolkit that to our knowledge contains the first end-to-end software pipeline for producing localized expression coefficients from a data-driven coding system.
\end{itemize}

\section{Related Work}

\subsection{Automated Action Unit Detection}
\label{sec:AU}

%

Automated FACS coding has long been a subject of intense research in computer vision~\cite{lien98, Essa97, bartlett99}, as it can support a variety of industrial and research applications~\cite{sariyanidi14}. However, there is still a need for improving accuracy, as the average F1 score of even state-of-the-art AU detectors is in the range of 0.60-0.67~\cite{chang22,yin24,ma24,liu24,hinduja23,baltrusaitis16}. The accuracy on cross-database experiments, which are indicative of real-world performance, tends to be even lower and below the threshold needed for reliable behavioral research~\cite{ertugrul20}. In an era where AI algorithms deliver impressive results across domains, the limited progress in automating FACS can be attributed to four structural barriers.

First, improving accuracy by increasing the training data is difficult. Training supervised models necessitates manual AU annotation, which is laborious and requires multiple trained experts. Even when longer video recordings of naturalistic social interactions are available, researchers are usually restricted to use only a few minutes or just seconds long segments (\eg most facially-expressive 20 seconds~\cite{zhang16}) due to the infeasibility of longer annotations. Second, AU labels are usable only if they pass a certain level of inter-rater reliability~\cite{ekman78} which can be low for certain AUs~\cite{zhang16,mcduff13}, reducing the usable video data further. Third, reliability tends to drift over time and between independent coders. Fourth, AUs that have a low base rate are difficult to detect with a supervised classifier may be excluded from automated pipelines altogether~\cite{hinduja23}. For example, existing toolkits provide outputs for only 17-19 AUs~\cite{baltrusaitis16,hinduja23,cheong23}, whereas the original FACS contains 45 AUs (30 for the revised version)~\cite{ekman02}. Finally, the AUs are generally not additive, as they can modify each other's appearance. A typical example is AU 1+4~\cite{cohn06}. When AU 1 occurs alone, the inner eyebrows are pulled upward. When AU 4 occurs alone, they are pulled together and downward. When AU 1 and AU 4 occur together, they result in appearance changes that do not occur in either AU 1 or AU 4 in isolation: The inner eyebrows are raised and pulled together, giving the brows an oblique shape, and causing wrinkles to appear in the center of the forehead. The existence of non-additive AUs suggest that automated FACS coders can comprehensively encode all expressions only if they are trained with datasets that contain all isolated AUs as well as non-additive AU combinations, which practically is not possible. 

While FACS is a powerful coding system, the above-listed barriers significantly restrict the upper limit of accuracy achievable by automated FACS coders. Our study focuses on developing a new coding system that retains a key property of FACS \--- breaking down expressions into localized units \--- while being more amenable to automation.

\subsection{Coding Systems Based on Linear Models}
\label{sec:rev_linear}

Linear decomposition based on sparse dictionary learning~\cite{mairal10} provides an alternative coding system that can readily overcome a fundamental limitation of FACS, namely, the lack of additivity~(Section~\ref{sec:FacialBasis}). In particular, sparse dictionaries can be trained to contain elements that correspond to localized expression units similar to AUs. An implementation of this idea for facial expression analysis is the Facial Bases method~\cite{sariyanidi17}. However, this method is based on a 2D pixel representation where separating out-of-plane head motion from facial expressions is not feasible. For this reason, producing meaningful expression coefficients from interactions that involve head movements (\eg conversations) is very challenging. Moreover, the Facial Bases method has not been compared to automated FACS in a clinical context, thus its ability to serve as an alternative coding system for mental health research is yet to be shown.

A potential alternative is to use 3D morphable model (3DMM) fitting~\cite{survey1} for reconstructing the 3D face shape from 2D data, as expressions can be separated from head pose and identity in the 3D coordinate space~\cite{sariyanidi24}. A 3DMM typically contains an expression model (\eg FaceWarehouse~\cite{cao13}) that, in principle, may serve as a coding system. The expression models of 3DMMs, however, typically contain deformations that can be physically implausible and impossible to interpret, as they are often generated using global models such as PCA~(Fig.~\ref{fig:PCA_components}; see Section~\ref{sec:FacialBasis}). In this paper, we define a sparse dictionary learning procedure that learns localized 3D facial expression units to generate physically plausible and interpretable movements akin to AUs~(Section~\ref{sec:exp_components}).

Similar interpretable and local linear models which operate in the 3D coordinate space have previously been used in the animation industry, where they are usually referred to as blendshapes~\cite{anjyo18}. The focus of blendshapes has been on video synthesis (3D to 2D) rather than analysis via reconstruction from videos. The number and content of the expression units in the blendshapes is determined according to this priority, which is not necessarily in line with the priorities of expression analysis. For example, the inclusion of as many as 946 expression units~\cite{anjyo18} may be warranted to generate person-specific differences in the appearance of expressions. Behavioral analysis, on the other hand, often demands a level of abstraction that ignores identity- or age-related differences in the generation of expressions. Representing facial behavior with hundreds of expression units can create multiple comparisons problems~\cite{curran00} and lead to multi-determined expression quantifications (same expression being represented by different components). As the distinctions between different expression units become too nuanced and difficult to semantically describe, reproducibility of quantitative findings becomes less attainable.

\subsection{Predicting Autism from Conversations}
Social communication is a core domain of impairment in autism~\cite{lord2012}. A number of studies aimed to delineate communication differences between autistic and neurotypical participants using video recordings of conversations and computer vision tools, showing that the two groups can be successfully classified~\cite{koehler24,zhang22,sariyanidi23}. 

The most common approach adopted by autism researchers to quantify facial expressions has been automated AU detection, as FACS is an established coding system used for nearly five decades in behavioral sciences. In particular, OpenFace has been the most widely used software, used in a large number of autism studies~\cite{zampella20,koehler24,openface1,openface2,openface3,openface4,openface5,openface6,openface7}. Although FACS is a reasonable choice in behavioral and medical sciences due to its interpretability and its ability to represent any possible facial expression, it is important to note that automated FACS coding is not equivalent to manual FACS coding. Detection accuracy for certain AUs can be low in automated software, and many AUs are entirely excluded, making it impossible to encode expressions comprehensively (Section~\ref{sec:AU}). To our knowledge, we apply the first alternative coding system to predict autism from conversational videos. Experiments show that the proposed system outperforms OpenFace in predicting autism, providing a promising alternative for studying mental health conditions. 

A common paradigm in the literature involves unstructured or semi-structured conversations, where participants engage in face-to-face, in-person interactions with study staff~\cite{sariyanidi23}. The COVID-19 pandemic accelerated the use of remote videoconferencing for telehealth applications, including autism assessment, due to their scalability, ecological validity, and convenience~\cite{de23}. It is therefore increasingly important to determine whether remote data collection paradigms can be effective in studying autism-related differences in social communication. To our knowledge, this is the first study that simultaneously conducts controlled AUT vs.\ NT classification using automated expression detection on data collected from the same paradigm from both in-person and remote conversations (Section~\ref{sec:datasets}). Results lead to two novel findings. First, in-person and remote conversations contain measurable differences in facial behavior between AUT and NT individuals, evidenced by a high and comparable classification accuracy (Section~\ref{sec:results}). Second, the behaviors that lead to highest classification accuracy are not necessarily identical between the two contexts.

\section{Proposed Expression Coding System}

The proposed coding system, Facial Basis, is a linear model that is comprised of localized expression components called the \textit{Basis Units} (BUs). Each BU represents a localized movement in the 3D coordinate space. This coding system provides two advantages. First, a facial expression is represented as the linear sum of BUs; therefore, the coding system does not suffer from the existence of non-additive units. Second, operating in the 3D rather than the 2D space allows facial expressions to be separated from head pose- or identity-related variations in an image.

Below we describe how we reconstruct the 3D face shape and expression variation in a given 2D image via 3DMMs~(Section~\ref{sec:3DMM}), and how we construct a coding system that represents a 3D expression variation as a linear sum of localized and semantically interpretable expression components~(Section~\ref{sec:FacialBasis}).

\subsection{Recovering Expression via 3DMM Fitting}
\label{sec:3DMM}

Let $\mathbf X \in \mathbb{R}^{3P}$ be a vector that represents the 3D shape of a face that appears in a given image\----a vector containing the 3D coordinates of $P$ points \textit{w.r.t.} the camera. Then, 3DMM fitting aims to reconstruct $\mathbf X$ as~\cite{sariyanidi24}
\begin{equation}
    \mathbf{X} = \mathbf{R(\bar {\mathbf X} + A\boldsymbol{\alpha} + E\boldsymbol{\varepsilon} )}+\boldsymbol{\tau},
    \label{eq:3DMM}
\end{equation}
where $\mathbf{\bar X}$ is the mean face of the 3DMM, and $\mathbf A$ and $\mathbf E$ are matrices that respectively represent the identity and expression models of the 3DMM. $\mathbf{R}$ and $\boldsymbol{\tau}$ account for the pose (rotation and translation) of the 3D face \textit{w.r.t.} the camera, whereas the vectors $\boldsymbol{\alpha}$ and $\boldsymbol{\varepsilon}$ respectively represent the identity- and expression-related variation in the 3D face shape. The decoupling of facial expression from pose and identity as in the right-hand-side of~\eqref{eq:3DMM} allows one to overcome two major challenges in expression analysis, namely, sensitivity to person-specific facial morphology (\ie identity bias) and to head pose~\cite{sariyanidi14}. It must be noted, however, that the decoupling of expression from pose or identity is an active research problem, and the degree to which it can be accurately accomplished depends on a number of factors, including the availability of the camera parameters~\cite{sariyanidi20a} or to the usage of single or multiple frames while reconstructing identity~\cite{sariyanidi24}, as well as the 3DMM fitting technique that is used. Nevertheless, the coding system that we propose is not tied to a specific 3DMM fitting procedure; as long as the mesh topology that underlies the 3DMM is compatible, the same coding system can be used with any fitting procedure. This modular design ensures the relevance of the solution to future advances in methods for characterizing expression, pose and identity. 

\subsection{Localized Facial Basis}
\label{sec:FacialBasis}

\begin{figure}
    \centering
    \includegraphics[width=0.95\linewidth]{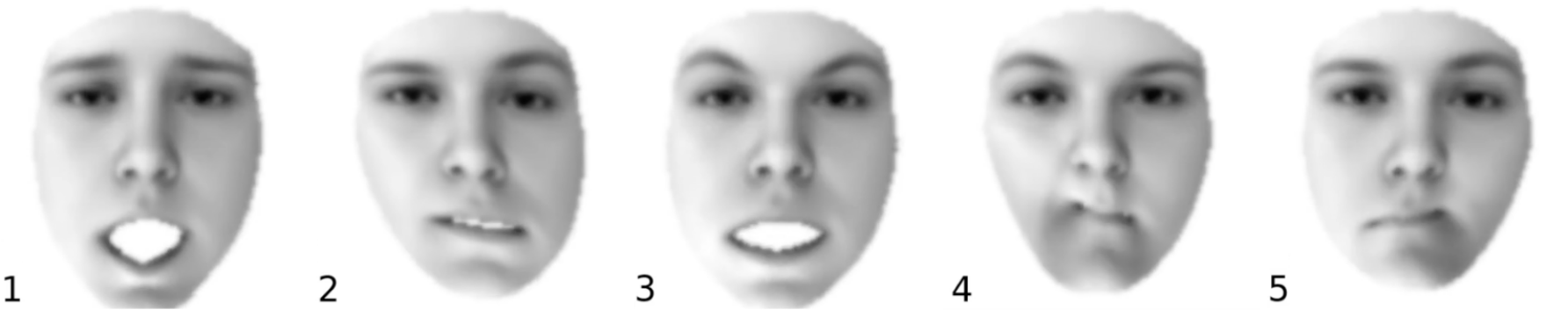}
    \caption{The first five components of the expression model used for 3DMM fitting~\cite{cao13}. Because this model is based on PCA, the components apply to the entire face and may be physically implausible.}
    \label{fig:PCA_components}
\end{figure}

Once a 3DMM is fit to an image, the facial expression in the image can be represented as $\mathbf{E}\boldsymbol{\varepsilon}$ in \eqref{eq:3DMM}, which is a $3P$-dimensional vector that describes how the 3D face shape deviates from the neutral face of the person. The matrix $\mathbf E$ can be considered as an expression coding system, since it explains the expression variation as a linear sum of its components (\ie columns) $\mathbf e_1, \mathbf e_2, \dots, \mathbf e_M$ as,
\begin{equation}
\mathbf{E}\boldsymbol{\varepsilon} = \varepsilon_1 \mathbf{e}_1  + \varepsilon_2 \mathbf{e}_2  + \dots + \varepsilon_{M}\mathbf{e}_{M} .
\end{equation}
That is, each of the components $\mathbf e_i$ can be considered as an expression code (\ie unit), and the coefficient $\varepsilon_i$ indicates whether the expression variation encoded by $\mathbf e_i$ is present in the input image, while the magnitude of $\varepsilon_i$ quantifies its intensity. However, the components $\{\mathbf e_i\}_i$ typically correspond to deformations that govern the entire face and can be physically implausible~(Fig.~\ref{fig:PCA_components}), as they are learned with PCA or a similar global transformation. Moreover, usually all the expression coefficients $\{\varepsilon_i\}_{i}$ are activated for any image that contains an expression variation. These characteristics are in contrast with FACS, which can represent any expression with a small subset of localized and specialized AUs. For example, the prototypical expressions of the six basic emotions involve 2-7 AUs~\cite{ekman97}. Indeed, the sparse activation pattern of FACS is a critical property for behavioral research, as it allows scientists to investigate which facial movements are related with an emotion, personality trait, or a mental condition.

Our coding system aims to reconstruct the expression  $\mathbf{E}\boldsymbol{\varepsilon}$ with another linear model $\mathbf W$ that emulates two properties of FACS: Containing localized expression components (\ie columns), and having a sparse activation pattern. Thus, when $\mathbf W$ is used to describe a facial expression as
\begin{equation}
    \mathbf{E}\boldsymbol{\varepsilon} \approx \mathbf {Wz} = z_1 \mathbf{w}_1 + z_2 \mathbf{w}_2 + \dots + z_K \mathbf{w}_K,
    \label{eq:W}
\end{equation}
only a subset of the coefficients $\{z_i\}_i$ are expected to be non-zero. 
While some of the expressions encoded in the components $\mathbf{w}_i$ can resemble AUs, in general, the coding system must be different from the FACS; otherwise, the non-additive AU combinations could not be reconstructed via a linear sum as in~\eqref{eq:W}. Indeed, experiments show that some non-additive AU combinations receive a dedicated component (Section~\ref{sec:exp_components}), although this does not imply that all non-additive combinations will receive a dedicated component. The exact nature of the components is determined by the procedure \---the algorithms and the data\--- that is used while learning the coding system. 

\subsection{The Procedure for Learning the Localized Face Basis}

We construct the model $\mathbf W$ from a dataset of $N$ images with facial expression variations, in an unsupervised manner. As a first step, we perform 3DMM fitting on every image, and obtain the corresponding vectors of expression coefficients $\boldsymbol{\varepsilon}_1, \boldsymbol{\varepsilon}_2, \dots, \boldsymbol{\varepsilon}_N$. Then, we learn the model $\mathbf W$ by fitting a sparse dictionary~\cite{mairal10}, which amounts to solving an optimization problem where the objective is to minimize the discrepancy between the two representations of the same expressions \---the representation via the 3DMM's expression model $\mathbf E$ and the targeted local basis $\mathbf W$\---, 
\begin{equation}
    \sum_n\|\mathbf{E}\boldsymbol{\varepsilon}_n - \mathbf {Wz}_n\|_2 + \lambda \sum_n \|\mathbf z_n\|_1, 
    \label{eq:obj}
\end{equation}
\textit{w.r.t.} $\mathbf W$ and $\{\mathbf z_n\}$. During optimization, we also enforce the $\ell_2$ norm of the components $\mathbf w_i$ to not exceed 1, otherwise they can grow unboundedly and compromise the desired sparsity pattern by allowing very small but non-zero values in $\mathbf z_n$ to have a significant impact~\cite{mairal10}.

The $\ell_1$ penalty $\lambda \sum_n \|\mathbf z_n\|_1 $ in \eqref{eq:obj} ensures that the representation via $\mathbf W$ will have a sparse activation pattern~\cite{mairal10}, satisfying one of the two properties of FACS that we try to emulate~(Section~\ref{sec:FacialBasis}). The other property that we are after, \ie having localized expression units, is not guaranteed by the terms in \eqref{eq:obj}. One way to achieve this property is adding an $\ell_1$ penalty on the expression units $\mathbf w_i$. While this practice is known to lead to spatially localized components~\cite{agarwal04,kim10,sariyanidi17} , there is no guarantee that all components will be localized\----the resulting components may span the entire input~\cite{sariyanidi17}. The existence of such global expression units is against a key property of FACS that we try to emulate. Thus, we enforce a constraint that results in a set of expression units where every unit is localized. This constraint is based on using the landmarks corresponding to facial features, as explained below. 

3DMMs typically contain $L$ indices that correspond to a subset of the $P$ mesh points that enclose the main facial features, namely the eyes, the brows, the nose and the mouth. For example, the iBUG-51 template~\cite{sariyanidi20a} contains $L{=}51$ landmarks. We use these landmarks to ensure that every expression unit is localized. This is achieved by allocating each expression unit $\mathbf w_i$ to a specific facial feature a priori, and allowing it to contain movement along the landmark points of the corresponding facial feature but not the other landmarks. For example, suppose that the left brow is represented with $L_{\text{LB}}$ landmark points. Then, each component $\mathbf{w}_i$ contains $3L_{\text{LB}}$ entries (\ie the 3D coordinates of $L_{\text{LB}}$ points) that correspond to the left brow landmarks. If a specific $\mathbf{w}_i$ is allocated to the left brow, then the corresponding $3L_{\text{LB}}$  entries are allowed to be non-zero, but the entries corresponding to the other $3(L-L_{\text{LB}})$ landmarks are forced to be zero. In this way, all the $K$ expression units are divided into six groups a priori, corresponding to the six facial features (two brows, two eyes, nose, mouth). Experiments show that the components learned in this way contain expression units that are localized\----they control predominantly one facial feature~(Section~\ref{sec:exp_components}).

\section{Experiments}

We first investigate the learnt Facial BUs and compare them to FACS AUs. Next, we show the utility of the proposed coding system in a clinical application by reporting results of classification experiments (AUT vs.\ NT) from conversational videos of adolescents. We provide comparisons with the most widely used automated FACS coder, namely, OpenFace~\cite{baltrusaitis16}. 

\subsection{Datasets}
\label{sec:datasets}
\textbf{Dictionary Learning.} We learned the Facial Basis by using the CK+~\cite{lucey10} and MMI~\cite{pantic05} datasets. While spontaneous datasets such as DISFA~\cite{mavadati13} or BP4D+~\cite{zhang16} can also be used, one must remove the video parts without facial movements and stratify the facial expressions before doing so, to ensure that the underlying reconstruction algorithm is not biased by the video parts without expressions and that expressions with low as well as high base rate are covered. The CK+ and MMI datasets are readily cropped to facial movements and contain a large variety of expressions, eliminating the need for preprocessing steps that can affect the learned expression components.

\textbf{Clinical Application.} We conducted clinical classification experiments on two datasets to assess the reproducibility of our findings in two different contexts, namely in-person face-to-face (F2F) conversations and remote room-to-room (R2R) conversations. The second context included videos collected through a lagless, cable-connected setup mimicking online video conferencing with ideal connectivity conditions between two separate rooms. English speaking participants included adolescents (age 12-17) drawn from a larger sample who participated in studies at The Children's Hospital of Philadelphia. The data collection procedure was approved by the Institutional Review Board (IRB) of The Children's Hospital of Philadelphia. The F2F dataset included 42 participants (AUT n = 21 [13 male]; NT n = 21 [13 male]), and the R2R dataset included 97 participants (AUT n = 49 [30 male]; NT n = 48 [29 male]). For both datasets, groups were matched on age, sex ratio, and intelligence quotient (IQ) (Table~\ref{tab:samples}). Autism diagnoses were confirmed through the best clinical judgment of a licensed psychologist using all available information, including administration of the Autism Diagnostic Observation Schedule (ADOS-2) \cite{lord2012}. NT participants did not have any history of mental health diagnosis per self- or parent-report.

Participants completed the Contextual Assessment of Social Skills \cite{Ratto2011}, a brief 3-4 minute semi-structured “get-to-know-you” conversation with a member of the research staff (confederates). Confederates were research assistants or students from the lab, assigned based on availability, whom the participant had not previously met. Participants and confederates were seated across from each other with two video cameras placed in between to record synchronized frontal videos of each person at 30 frames per second. Confederates were instructed to appear interested and engaged but not carry the conversation (\ie speak no more than 50\% of the time and wait 5 seconds to re-initiate the conversation after a lapse).


\begin{table}[]
    \centering
        \caption{The (mean) age and IQ; and number of female and male participants in the two study samples. $p$ values indicate possible group differences.}
    \begin{tabular}{l|ccc|ccc}\hline
    & \multicolumn{3}{c|}{Face-to-Face (F2F)} & \multicolumn{3}{c}{Room-to-room (R2R)} \\
   &  	AUT & 	NT	& $p$ val. & AUT & 	NT	& $p$ val. \\ \hline\hline
Age &	14.3  & 14.2 & 0.90 & 15.2  & 15.1 & 0.78 \\
IQ	 & 98.1 &	108.0 &	0.06 &  113.0 &	112.0 &	0.78 \\
F/M participants &	8/13 &  8/13 &	1.00 & 19/30 &	19/29 &	1.00 \\ \hline
    \end{tabular}
    \label{tab:samples}
\end{table}

\subsection{Experimental Setup}
\label{sec:exp_setup}


\textbf{Compared coding systems.} We compare the prediction accuracy of four facial expression coding systems: FACS coding via OpenFace, Facial Basis (Section~\ref{sec:FacialBasis}), PCA (Section~\ref{sec:3DMM}) and local PCA. The latter contains feature-specific expression components like the BUs, but the components are still learned using PCA instead of sparse coding. For each coding system, we investigate how performance varies with the number of components used. Specifically, we report classification accuracy obtained using the first $k$ components, where $k$ is increased with increments of five (\ie $k=5, 10, \dots, K_\text{exp}$). For these experiments, the expression components are ordered according to their magnitude of activation on the datasets used for the experiments. That is, the $k$th expression component of a coding system is the one that had the $k$th highest magnitude across the F2F and R2R datasets.

\textbf{Classification pipeline.} Given the restrictions that stem from the clinical sample sizes, we conducted experiments using shallow classification pipelines. Also, to reliably tune the pipelines through (nested) cross validation, we aimed to minimize the number of hyperparameters and we used a linear SVM classifier, which contains only one parameter to be tuned, namely, the $C$ parameter. The raw input data are defined by $K$ facial expression signals and $3$ head movement signals (\ie rotation angles). The head movements are included in FACS as facial action descriptors~\cite{ekman78} and are known to carry meaningful communication cues that vary with mental health conditions~\cite{horigome20,mcdonald23}. The facial expression signals at every frame are obtained either using a data-driven coding system or OpenFace. The final input features are generated from these $Q = K+3$ signals using (intra-person) windowed cross-correlation (WCC). WCC is a widely used approach in behavioral research~\cite{boker02} and can capture behavioral differences that are expected to exist between AUT and NT groups, such as typicality/atypicality of expressions~\cite{brewer16} or level of integrating multiple components of behavior~\cite{hus14}. WCC represents the behavior within a time window of $T_w$ seconds through a vector of $Q^2$ features (all possible pairs of signals). The behavior over the entire video is represented as a $Q^2$-dimensional vector obtained by averaging over the feature vectors across all time windows within a video~\cite{sariyanidi23}. These $Q^2$-dimensional feature vectors are used with the SVM classifier. Results are reported with leave-one-out cross-validation, and the $C$ parameter at each cross-validation fold is tuned with a 5-fold inner cross-validation (\ie nested cross validation).

\textbf{3DMM fitting.} The 3DMM fitting procedure needed for the Facial Basis~(Section~\ref{sec:3DMM}) was devised to be computationally efficient while also representing expressions comprehensively. Existing deep learning-based 3DMM fitting software are usually computationally efficient but represent expressions with reduced capacity. For example, the methods based on the Basel Face Model (BFM)~\cite{paysan09} topology may use only 29~\cite{chang17,guo20} of the $M=79$ components of associated expression model~\cite{cao13}. Optimization-based methods such as 3DI~\cite{sariyanidi24}, on the other hand, can use all expression components as they do not commit to a specific version of a 3DMM, but are prohibitively slow or require GPU during inference. Thus, we trained a ResNet, which predicts identity, pose and expression parameters (Section~\ref{sec:3DMM}) according to the BFM model from an image frame in a supervised fashion, where the labels are obtained by using the 3DI method. Specifically, we trained using a large dataset (\ie  YouTube Faces~\cite{wolf11} and CelebA~\cite{liu18} datasets combined) to minimize the L2 loss between the labels predicted by the ResNet and the labels obtained by 3DI prior to training.

\textbf{Implementation details.} The Facial Basis was learned using \texttt{scikit-learn}, which implements the minimization of \eqref{eq:obj} with its dictionary learning module. We set the $\lambda$ coefficient of~\eqref{eq:obj} to 0.2, and the number of components of the coding system $\mathbf W$ to $K{=}50$, as qualitative inspection suggested that increasing $K$ further led to basis components that are highly similar. Each expression component in $\{\mathbf w_i\}_{i=1}^K$ corresponds to a specific facial feature, and for convenience, we associated a two-letter code to each of the components according the feature they correspond to (LB: left brow, RB: right brow, LE: left eye, RE: right eye, NO: nose and MO: mouth). Thus, each component $\mathbf w_i$ has a unique name, constructed by using the two-letter code and a number. For example, LE-3 is the third expression component associated with the left eye.

\subsection{Results}

\subsubsection{The Learned Expression Components}
\label{sec:exp_components}
\begin{figure}
    \centering
    \includegraphics[width=1.0\linewidth]{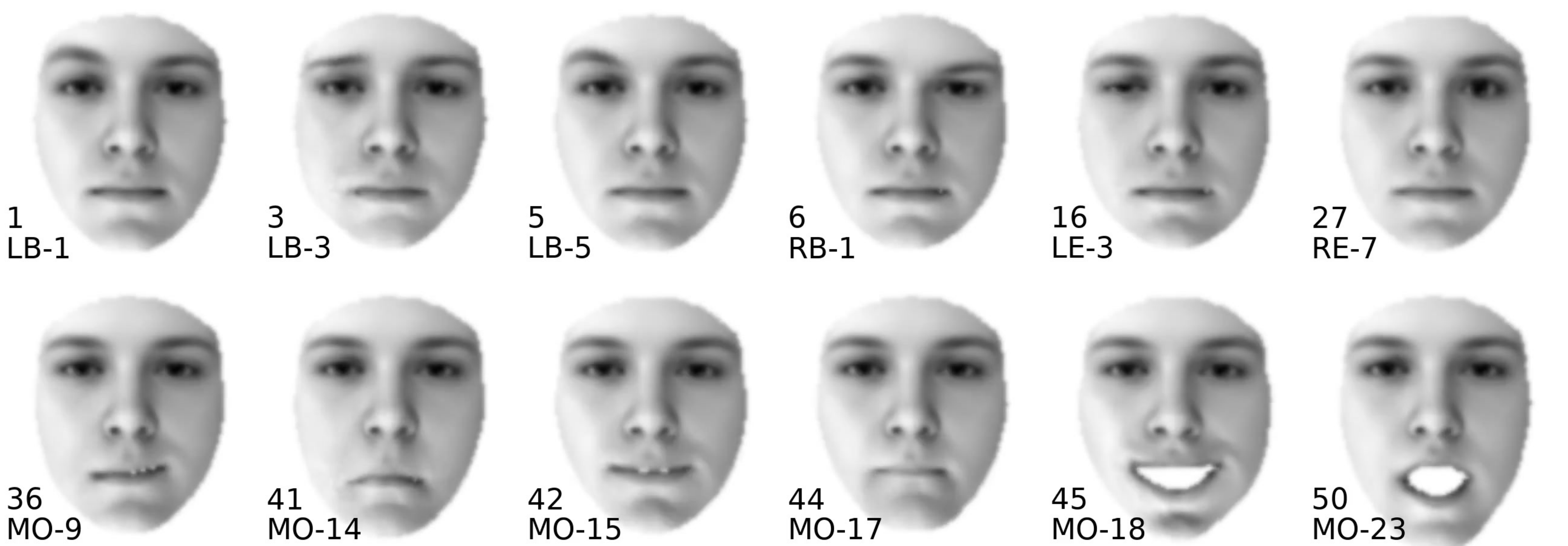}
    \caption{The expression encoded by some Facial Basis Units (BUs).}
    \label{fig:basis}
\end{figure}

Fig.~\ref{fig:basis} shows the expressions encoded by some of the FBUs, and suggests that many components correspond to plausible and interpretable facial movements\----movements that are localized and can be generated by the human face. This is in contrast with the PCA-based expression basis of the 3DMMs, which contain expression components that are not localized and can be physically implausible (Fig.~\ref{fig:PCA_components}). The BUs show some visible similarities to FACS AUs, but they also diverge from them. 

\begin{figure*}
    \centering
    \includegraphics[width=1.0\linewidth]{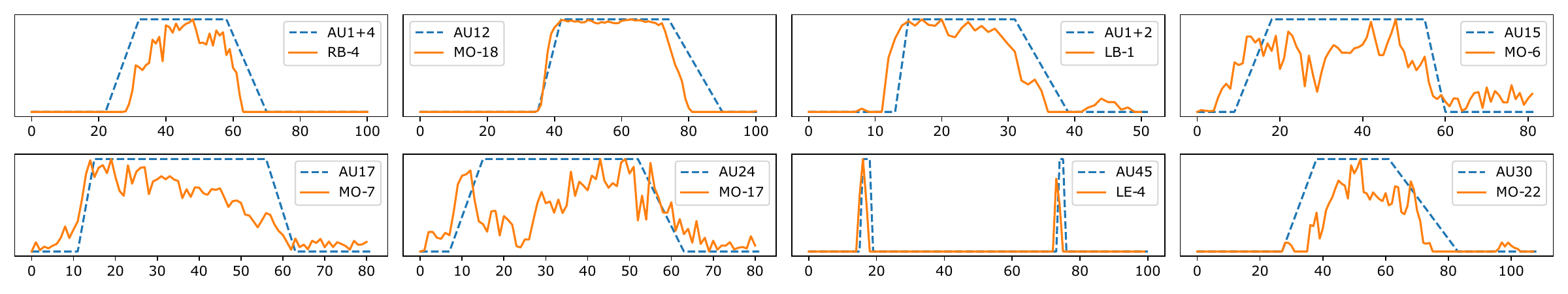}
    \caption{The AU labels (ground truth) from videos of the MMI dataset and the BU coefficients of the corresponding expression units, plotted over time. Results suggest that both the FACS AUs and the Facial BUs can be used to infer localized movements. }
    \label{fig:MMI_examples}
\end{figure*}

The existence of differences between these two coding systems is expected by design, since our goal is not to replicate FACS but create an alternative coding system that can represent expressions as a linear sum, which cannot be done by FACS due to the non-additivity of some AUs~(Section~\ref{sec:FacialBasis}). For example, one of the Facial BUs resembles AU 1+4 (LB-3 in Fig.~\ref{fig:basis}), which is a reasonable outcome, as AU 1 and AU~4 are not additive, and a linear model that contains only these units would not be able to represent their combination. Table~\ref{tab:AU_labels} shows more examples of similar-looking BUs and AUs (or AU combinations). Fig.~\ref{fig:MMI_examples} shows that despite the lack of one-to-one matching between FACS AUs and the Facial BUs, both systems can be used to infer localized movements in facial videos. The visualization of  all 50 BUs is provided at \url{github.com/sariyanidi/FacialBasis}.

A critical advantage of the Facial Basis compared to automated FACS software is that it readily supports the analysis of asymmetric expressions (see modifiers R and L in Table~\ref{tab:AU_labels}). Automated FACS software outputs typically do not provide predictions for AUs that occur only on one hemiface~\cite{baltrusaitis16,hinduja23,cheong23}, in part due to the difficulties related to eliciting asymmetric facial expressions. Being able to capture asymmetrical expressions or quantifying the degree of symmetry can be important for predicting autism~\cite{guha15} or studying neurological conditions such as cerebral palsy.

\begin{table}[]
    \centering
        \caption{Left: Correspondences between Facial Basis Units (BUs) and AUs. Right: Description of specified AUs. Following FACS annotation, asymmetric AUs are denoted by `L' (if only on the left hemiface) or `R' (if only on the right hemiface).}
    \begin{tabular}{|l|l|} \hline
    \makecell{Facial Basis\\Unit (BU)} & \makecell{AU\\Combination} \\ \hline
LB-1	&	AU L1+L2	\\ 
LB-3	&	AU L1+L4	\\
LB-5	&	AU L2	\\
RB-1	&	AU R4	\\
LE-3	&	AU 41	\\
RE-7	&	AU 5	\\
MO-9	&	AU R12+R14A	\\
MO-14	&	AU 15	\\
MO-15	&	AU 12	\\
MO-17	&	AU 24	\\
MO-18	&	AU 12+27	\\
MO-23	&	AU 26	\\    \hline
\end{tabular}\quad\quad
\begin{tabular}{|l|p{2.13cm}|} \hline
\makecell{AU\\Code} & Description  \\ \hline
AU 1	&	Inner Brow Raiser	\\
AU 2	&	Outer Brow Raiser	\\
AU 4	&	Brow Lowerer	\\
AU 5	&	Upper Lid Raiser	\\
AU 12	&	Lip Corner Puller	\\
AU 14	&	Dimpler	\\
\makecell[l]{AU 15\\ { }}	&	\makecell[l]{Lip Corner \\Depressor}	\\
AU 24	&	Lip Pressor	\\
AU 26	&	Jaw Drop	\\
AU 27	&	Mouth Stretch	\\    
AU 41	&	Lid Droop	\\    \hline
\end{tabular}
    \label{tab:AU_labels}
\end{table}

\begin{figure}
    \centering
    \includegraphics[width=0.66\linewidth]{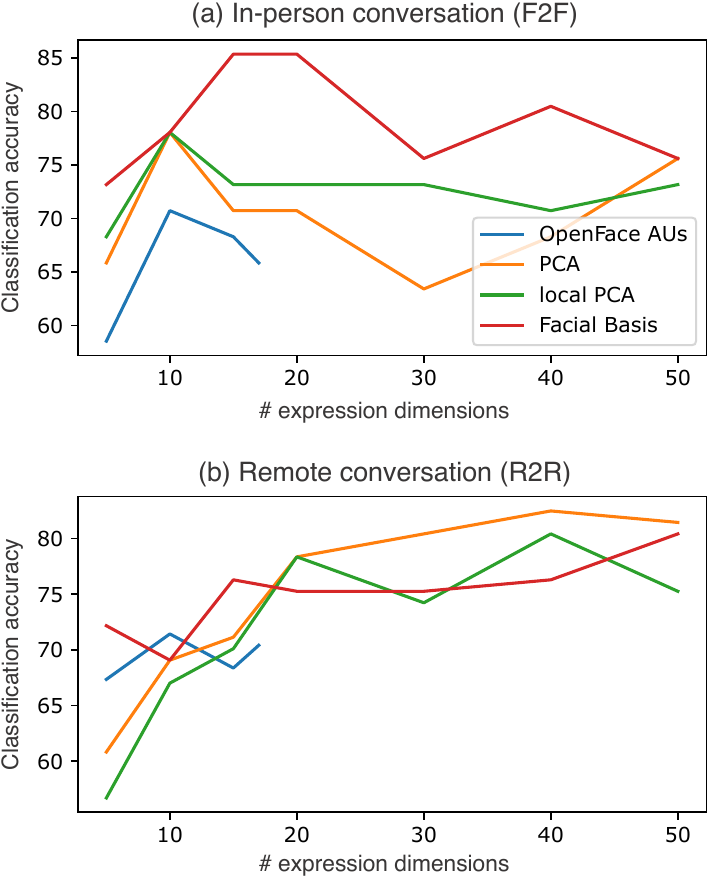}
    \caption{AUT vs. NT classification results of the compared coding systems w.r.t. the number of expression components used per coding system.}
    \label{fig:results}
\end{figure}

\subsubsection{Clinical Classification Results}
\label{sec:results}
Fig.~\ref{fig:results} reports the AUT vs.\ NT classification performance of the compared coding systems, and shows how performance varies with the number of expression components used. On the F2F sample, the highest classification accuracy is achieved by the Facial Basis (Fig.~\ref{fig:results}a). Of note, all data-driven coding systems outperform OpenFace. All methods reach their peak performance with 10-15 expression components on the F2F sample, suggesting that a relatively small number of expression units may suffice to achieve peak accuracy. However, the trend is different for the R2R sample. All methods other than automated FACS reach their peak performance around 40-50 units (Fig.~\ref{fig:results}b). The number of components examined for FACS was limited to 17, the maximum number of AU intensity estimates provided by OpenFace.

The different trends between F2F (Fig.~\ref{fig:results}a) and R2R (Fig.~\ref{fig:results}b) suggest that behavioral characteristics of in-person communication may be different from those of a remote, computer-based communication, even when the latter is lagless. To further investigate potential differences caused by the communication medium, we performed classification experiments using only head movements. On the F2F sample, the classification accuracy of head movements alone was 73\%, which is consistent with previous literature suggesting that head movements contain rich information to classify between the AUT and NT groups~\cite{martin18,mcdonald23,gokmen24}. However, the classification accuracy on the R2R sample was only 32\%, providing further evidence about the existence of behavioral differences between in-person vs.\ remote conversations.

\begin{figure}
    \centering
    \includegraphics[width=0.9\linewidth]{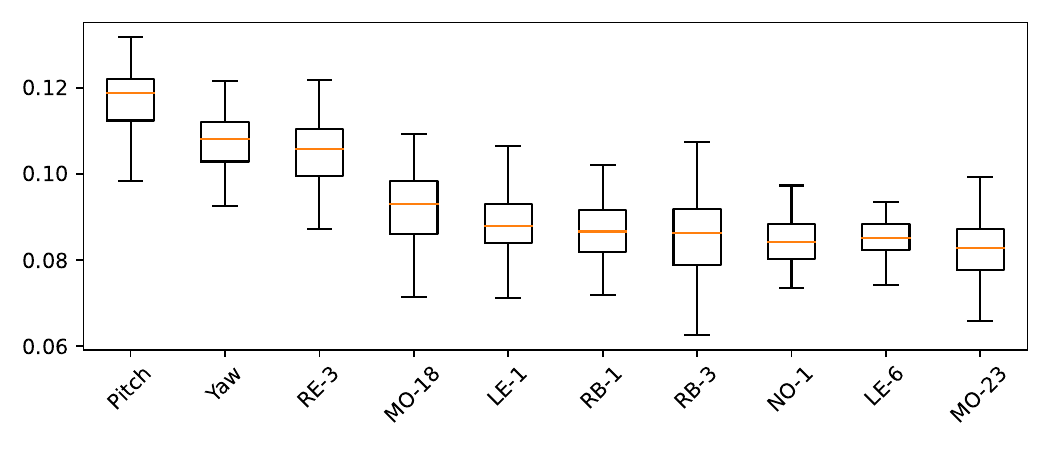}
    \includegraphics[width=0.9\linewidth]{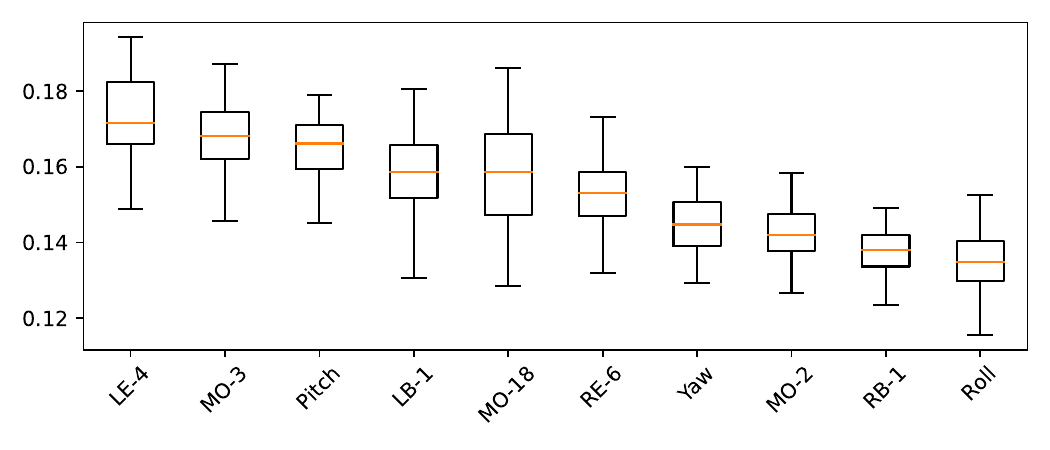}
    \caption{Average feature weights of behavioral components that include head movement (Pitch, Yaw, Roll angles) as well as Facial BUs. Top: in-person sample (F2F); bottom: remote (R2R) sample.}
    \label{fig:feature_weights}
\end{figure}

Fig.~\ref{fig:feature_weights} provides further insights about the differences between the two recording setups by illustrating the ten Facial BUs that received the highest weight by the SVM classifiers. Since feature weights depend on the training sample, we plot a distribution (\ie boxplot) per component, constructed by using the feature weights of the corresponding component across 100 randomly picked subsamples of the training data. The head movement components (pitch and yaw rotation) hold the highest weight in the F2F sample but not in the R2R sample, further supporting the importance of head movements during in-person communication. The next most important feature in the F2F sample is RE-3, which corresponds to closing eyelids (see Supplementary Video). The high weight of this component may be attributed to atypical eye-blink patterns that have been observed in autism \cite{sears94,krishnappa23} or its indirect link to social gaze behavior~\cite{yankowitz24}. The feature with the next highest weight is MO-18, which corresponds to a smile with open mouth (Fig.~\ref{fig:basis}), and the importance of this feature may be explained by the atypicality of smile~\cite{metallinou13} in youth with autism or reduced inter-personal affect coordination \cite{zampella20} in autism. 

An important detail in Fig.~\ref{fig:feature_weights} is that the classifier weights for the R2R setup show slower decay, further supporting the need of a richer representation with an increased number of components. While in general the features that lead to the highest classification accuracy depend on the recording setup, MO-18 is one of the components that is in the top five of both recording setups, and another one is the pitch rotation, which occurs with head nodding. Another high-ranking feature in the R2R sample is MO-3, which includes a lip corner movement that can also be observed with a smile. It must be noted that the weight corresponding to each expression component in Fig.~\ref{fig:feature_weights} is depicted after averaging all the cross-correlation features that include the component. Therefore, the listed expression components are not necessarily important in isolation, but in combination with other components, since the cross-correlation features encode the relationship between different behavioral components~(Section~\ref{sec:exp_setup}). A more nuanced analysis would require the inspection of the individual features without averaging. However, such an analysis requires larger samples, as the weight of individual features had too much variance and instability in our sample, which is expected when the number of individual features is high~\cite{khaire22}, as in our pipeline ($Q^2{=}53^2{=}2809$).

In sum, our experiments provide three critical insights. First, data-driven coding outperforms the most widely used AU detector in predicting autism. Second, automated facial expression analysis in mental health may require or benefit from more expression coefficients than those provided by automated FACS coders. Third, in-person and remote communication likely have different behavioral dynamics.

\section{Limitations}
\label{sec:limitations}
The presented coding system suffers from four limitations that reduce its interpretability or representation power. First, while most Facial BUs are interpretable and physically plausible, there are a number of units that are difficult to semantically describe or unlikely to be generated by a face (\eg MO-4, MO-9 and MO-12 in supplementary video). These units can be reduced or eliminated by augmenting the objective function with loss terms that enforce expression units to be physically plausible by imposing certain anatomical constraints~\cite{ichim17}. The second limitation is that some of the expression units describe very similar movements, and the lack of distinct differences between expression units can complicate analyses (Section~\ref{sec:rev_linear}). This issue can be remedied by augmenting the objective function~\eqref{eq:obj} with terms to prevent BUs from being highly similar, or by creating a hierarchy, akin to hierarchical clustering~\cite{murtagh12}, where expression units with a high degree of similarity are clustered together. Third, our classification experiments are based on a spontaneous task, but the Facial Basis is currently trained with only posed expressions. To better represent naturalistic expressions, datasets collected from spontaneous tasks or interactions~\cite{mavadati13,zhang16} can be included. The inclusion of spontaneous videos may require additional pre-processing procedures or using alternative loss function (\eg wing loss~\cite{feng18}) to ensure that expressions with low as well as high base rates are represented by the learned coding system. 
Finally, the Facial BUs are limited in their ability to capture wrinkles and furrows that may occur with expressions. This is likely caused by the fact that the 3DMM that we use to model the 3D face, BFM, has a relatively low resolution, as it represents the facial region with approximately 20k points. This limitation may be remedied by the usage of 3DMMs with higher resolution, such as the FaceScape~\cite{yang20}, which represents 3D face with  approximately 2 million points.

A limitation on the interpretation of the differences between the R2R and F2F setups is that recordings were drawn from two independent samples. While these samples are demographically and clinically similar, it is possible that underlying differences in the participants in each sample drove differences, rather than recording setup.

\section{Discussion and Outlook}
\label{sec:discussion}
Our results suggest that a data-driven coding system provides an interpretable approach for encoding facial behavior, and outperforms a standard FACS coder in the scope of predicting autism. Thus, we show that a promising new paradigm is to develop alternative coding systems that mimic FACS rather than developing automated coders that mimic FACS experts, particularly because the accuracy of the latter approach may be saturating after more than forty years of research.

An immediate impact of the proposed study is to use the specific coding system, namely Facial Basis, for studying mental conditions and other behavioral research questions (\eg emotions). It should be noted, however, that the proposed system is simply a starting point and its current limitations (Section~\ref{sec:limitations}) suggest that its output must be interpreted with caution. The restricted dataset used during its training may not be representative across a variety of research questions (\eg detection of pain or prediction of emotional states).

The possibly more significant but longer-term impact of our study is the construction of a universal coding system similar to FACS that can serve as a common language for quantifying behavior across studies, contexts and mental health conditions. A system of this kind would encode facial expressions comprehensively, reliably, and validly, while also proving successful across a variety of mental health applications. As such, the construction of a universal coding system must be the result of more intense efforts spanning multiple studies in diverse populations. The dataset that is used for such a coding system must be large enough to cover virtually any facial action that can be generated by a face, while the learning procedure must ensure that expressions with low as well as high base rate are represented. 

\section*{Ethical Impact Statement}
The main objective of this study is to advance behavioral and mental health research by providing scientists with a tool that encodes facial behavior more comprehensively than existing tools. This tool, however, could be used by malevolent actors who want to improve skills and capabilities that are detrimental to society, such as deception with facial behavior.




{\small
\bibliographystyle{IEEEtran}
\bibliography{main_arxiv}
}

\end{document}